\documentclass[11pt]{article}
\usepackage[letterpaper,margin=1in]{geometry}

\usepackage[T1]{fontenc}
\usepackage{amsmath,amssymb,amsthm,mathtools,bm}
\usepackage{graphicx,booktabs,multirow,array,xcolor,microtype}
\usepackage[round,authoryear]{natbib}
\usepackage[colorlinks=true,linkcolor={blue!60!black},citecolor={blue!60!black},urlcolor={blue!60!black}]{hyperref}
\usepackage{xurl}
\usepackage{tikz}
\usetikzlibrary{positioning,arrows.meta,fit,backgrounds,shapes.geometric,calc,patterns}
\usepackage{float}
\usepackage{algorithm}
\usepackage{algpseudocode}
\definecolor{rgBlue}{HTML}{2F6690}\definecolor{rgBlueL}{HTML}{9BB8D3}\definecolor{rgBluePale}{HTML}{E9F0F7}
\definecolor{rgOrange}{HTML}{D99043}\definecolor{rgOrangeDk}{HTML}{9A5B1E}\definecolor{rgOrangePale}{HTML}{FBF0E3}
\definecolor{rgInk}{HTML}{25313B}\definecolor{rgGrayDk}{HTML}{5B6670}\definecolor{rgGray}{HTML}{8C96A0}
\definecolor{rgGrayL}{HTML}{C9CED4}\definecolor{rgGrayPale}{HTML}{F2F4F6}\definecolor{rgRed}{HTML}{C75C58}

\newtheorem{proposition}{Proposition}
\newtheorem{assumption}{Assumption}
\theoremstyle{remark}
\newcommand{\Dtr}{D_{\mathrm{tr}}}\newcommand{\Dsel}{D_{\mathrm{sel}}}
\newcommand{\Dcal}{D_{\mathrm{cal}}}\newcommand{\Dte}{D_{\mathrm{te}}}
\newcommand{\Dh}{\widehat{\Delta}}
\newcommand{\eprac}{\epsilon_{\mathrm{prac}}}
\newcommand{\pmatch}{p_{\mathrm{match}}}
\newcommand{\plegacy}{p_{\mathrm{legacy}}}
\newcommand{\Gc}{\mathcal G_{\mathrm{country}}}
\newcommand{\Gy}{\mathcal G_{\mathrm{year}}}

\hypersetup{pdftitle={RAEGL: Deploying Contextual Corrections over a Frozen Forecaster under Panel Dependence},pdfauthor={Yifan Guo}}
\begin{document}
\title{\textbf{RAEGL: Deploying Contextual Corrections over a\\Frozen Forecaster under Panel Dependence}}
\author{Yifan Guo\\[2pt]Stony Brook University}
\date{}
\maketitle

\begin{abstract}
A frozen global forecaster can often be adjusted by an additive correction indexed by categorical context such as region or income group. Deploying the correction is a decision with two errors: a false deployment worsens future forecasts, and a conservative miss forgoes an improvement. We study this decision with RAEGL, a protocol that separates training, candidate selection, independent calibration and sealed evaluation, and that combines a calibration-gain floor with a label-permutation screen and a per-block stability check. Three findings emerge. First, the permutation group determines whether the screen is valid. In 240 synthetic zero-signal panels with persistent country effects, permuting whole-country labels rejects in 4.2\% of panels, while shuffling labels within years rejects in 74.2\%; re-running selection inside each permutation changes little. Second, independent calibration accounts for the reduction in harm. Across 720 simulated deployments, the full rule lowers harmful deployments from 42.1\% to 18.5\% of trials, and a plain calibration-gain threshold reaches the same 18.5\% and agrees with it in 718 trials. Third, on World Development Indicators the locked rule falls back to the global forecaster; the rejected correction would have changed test RMSE by 0.17\%, with a bootstrap interval covering zero. A two-point argument shows that no calibration-based rule can detect a reversal that first appears in the test period.
\end{abstract}
\raggedbottom

\section{Introduction}
Forecasting systems often pool heterogeneous entities. Country-year panels combine economies with different structures, policies and reporting practices, and contextual models or mixtures of experts exploit this heterogeneity by allocating conditional capacity \citep{jacobs1991adaptive,shazeer2017outrageously,hazimeh2021dselect}. When a validated global predictor already exists, a contextual correction is an optional component. Deploying an unsupported correction can make future forecasts worse than keeping the global model, and refusing a useful one forgoes an improvement. Neither error is visible if contextual specialization is treated as an automatic consequence of model selection.

We treat activation of a contextual correction as one decision for a declared horizon: deploy the selected correction for every unit, or return the global forecast exactly. RAEGL (Risk-Aware Evidence-Gated Learning) makes this decision from data that the selection step never used. It requires the correction to clear a declared minimum gain on a calibration period, to pass a label-permutation screen, and to improve every calibration block. The protocol is simple; our interest is in what each guard contributes and under which assumptions it is valid. We ask three questions. Which label permutations give a valid screen when units carry persistent effects? Does the screen, or the block check, change decisions relative to a plain calibration threshold? What remains undetectable before deployment?

\textbf{Contributions.}
(i)~We formalize the deploy-or-fallback decision, its two error types and the difference between unconditional and conditional harm, and give an exact identity (Proposition~\ref{prop:increment}) showing that a guard added to a threshold rule improves regret only through the candidates it vetoes, and only if their future gain is negative.
(ii)~We show that under panel dependence the validity of the permutation screen depends on the permutation group far more than on the statistic. Standard random-permutation results (Propositions~\ref{prop:valid}--\ref{prop:legacy}) apply only when the group preserves the persistence of unit labels; for geographic labels in observational data this remains an assumption.
(iii)~On WDI, on 12 retrospective origin--residual cells and on 720 synthetic trials, the additional guards are almost always decision-redundant given the calibration threshold, and a two-point bound (Proposition~\ref{prop:reversal}) shows that reversals confined to the test period cannot be detected by any rule that sees only earlier data.

The practical recommendation is correspondingly modest: hold out an independent calibration period and require a declared minimum gain on it; add a permutation screen only when a dependence-preserving group can be justified, and do not expect it to reduce harm by itself.

\section{Related Work}\label{sec:related}
\textbf{Contextual forecasting and selective deployment.} Mixture-of-experts and sequence forecasters allocate context-dependent capacity \citep{jacobs1991adaptive,shazeer2017outrageously,hazimeh2021dselect,salinas2020deepar,lim2021tft}; RAEGL leaves the forecasting architecture unchanged. Selective prediction abstains per observation \citep{chow1970optimum,geifman2017selective}, whereas our action enables one component for a whole horizon. The closest decision analogue is safe policy improvement over a baseline with a high-confidence release criterion \citep{thomas2015hcpi,laroche2019spibb}.

\textbf{Forecast comparison, search and calibration.} Predictive-ability tests address nested models, temporal instability, panel data and multiple specifications \citep{diebold1995comparing,white2000reality,hansen2005spa,clark2007approximately,giacomini2006conditional,giacomini2010forecast,qu2024panel}. The calibration-maximum statistic used for our WDI lock belongs to the max-statistic tradition of \citet{white2000reality} and \citet{hansen2005spa}; the matched statistic instead reruns the selection step. Holdout randomization \citep{candes2018panning,tansey2022holdout} and exchangeability blocks \citep{winkler2015multilevel} clarify which symmetry a permutation test needs. Post-selection inference \citep{berk2013posi} and learn-then-test or risk control \citep{bates2021rcps,angelopoulos2025ltt,angelopoulos2024crc,zecchin2025altt} are natural alternatives; testing the calibration risk of the single frozen correction is a learn-then-test-style comparator, and we include a confidence-bound rule of this kind. Model selection under temporal shift \citep{han2024temporal}, temporal evaluation \citep{tashman2000out} and adaptive conformal methods \citep{gibbs2021aci,barber2023beyond} do not remove the need for stability between calibration and deployment.

\section{Problem Formulation and Protocol}\label{sec:method}
\begin{figure}[t]
\centering
\resizebox{\textwidth}{!}{
\begin{tikzpicture}[
  font=\scriptsize, >={Stealth[length=3.6pt,width=3.2pt]}, text=rgInk,
  box/.style={draw=rgGray, line width=.4pt, rounded corners=1.5pt, fill=white, align=center,
              minimum width=2.75cm, minimum height=.62cm, inner sep=2pt, text width=2.65cm},
  frozen/.style={box, draw=rgInk, line width=.65pt},
  gate/.style={box, draw=rgBlue, line width=.6pt, fill=rgBluePale},
  lane/.style={fill=rgGrayPale, draw=none, rounded corners=3pt},
  lanetitle/.style={font=\scriptsize\bfseries, text=rgInk, align=center},
  flow/.style={->, draw=rgGrayDk, line width=.5pt},
  bflow/.style={->, draw=rgBlue, line width=.65pt},
  dec/.style={diamond, aspect=1.55, draw=rgBlue, line width=.65pt, fill=white, align=center, inner sep=0.5pt, font=\scriptsize},
]
\def\xa{0} \def\xb{3.25} \def\xc{6.5} \def\xd{9.75} \def\xe{13.0}
\node[box] (a1) at (\xa,0) {Training-only\\preprocessing};
\node[box, below=.22cm of a1] (a2) {Fit global predictor $g$\\(WDI: Random Forest)};
\node[box, below=.22cm of a2] (a3) {Residuals $e{=}y{-}g(x)$;\\centered codes $\tilde\phi_j(c)$};
\node[box, below=.22cm of a3] (a4) {Ridge fits $r_{j,\lambda}$ for\\all $(j,\lambda)\in\mathcal J{\times}\Lambda$};
\node[frozen, below=.22cm of a4] (a5) {\textbf{Frozen:} $g$, prepro-\\cessing, grid $\mathcal J{\times}\Lambda$};
\node[box] (b1) at (\xb,-.84) {Score $\widehat\Delta_{\mathrm{sel}}(h_{j,\lambda};g)$\\for every declared pair};
\node[box, below=.22cm of b1] (b2) {$(j^\star\!,\lambda^\star){=}\arg\max\widehat\Delta_{\mathrm{sel}}$\\(\textsf{NoContext} scores 0)};
\node[frozen, below=.22cm of b2] (b3) {\textbf{Frozen:} $h^\star{=}h_{j^\star\!,\lambda^\star}$;\\eligible iff $\widehat\Delta_{\mathrm{sel}}{>}0$};
\node[gate] (c1) at (\xc,-.5) {\textbf{(i) Randomization}\\$p\le\alpha$ (group $\mathcal G$)};
\node[gate, below=.3cm of c1] (c2) {\textbf{(ii) Practical floor}\\$\widehat\Delta_{\mathrm{cal}}(h^\star;g){\ge}\epsilon_{\mathrm{prac}}$};
\node[gate, below=.3cm of c2] (c3) {\textbf{(iii) Temporal guard}\\$\min_{b\in\mathcal B}\widehat\Delta_{\mathrm{cal},b}{>}0$};
\node[dec] (d0) at (\xd,-1.75) {Eligible and\\(i)--(iii)\\all hold?};
\node[box, draw=rgBlue, line width=.75pt, fill=rgBluePale, above=.42cm of d0] (dy) {\textbf{Deploy} $h^\star$\\for the whole period};
\node[box, draw=rgInk, line width=.65pt, below=.42cm of d0] (dn) {\textbf{Exact fallback}\\$\hat y=g(x)$};
\node[box] (e1) at (\xe,0) {Open $D_{\mathrm{te}}$ once,\\only after locking};
\node[box, below=.22cm of e1] (e2) {Realized gain\\$a\cdot\widehat\Delta_{\mathrm{te}}(h^\star;g)$};
\node[box, below=.22cm of e2] (e3) {False deployment,\\miss, oracle regret};
\node[box, below=.22cm of e3] (e4) {Post-decision diagnostics\\(bootstrap, years, regions)};
\node[box, draw=rgGray, dashed, text=rgGrayDk, below=.22cm of e4] (e5) {No feedback\\to phases A--D};
\begin{scope}[on background layer]
 \node[lane, fit=(a1)(a5), inner sep=4pt] (LA) {};
 \node[lane, fit={(b1.west|-a1.north)(b1.east|-a5.south)}, inner sep=4pt] (LB) {};
 \node[lane, fit={(c1.west|-a1.north)(c1.east|-a5.south)}, inner sep=4pt] (LC) {};
 \node[lane, fit={(dy.west|-a1.north)(dy.east|-a5.south)}, inner sep=4pt] (LD) {};
 \node[lane, fit={(e1.west|-a1.north)(e1.east|-a5.south)}, inner sep=4pt] (LE) {};
\end{scope}
\node[lanetitle, above] at (\xa,0|-LA.north) {A.~Training\\{\normalfont\color{rgGrayDk}$D_{\mathrm{tr}}$}};
\node[lanetitle, above] at (\xb,0|-LA.north) {B.~Selection\\{\normalfont\color{rgGrayDk}$D_{\mathrm{sel}}$}};
\node[lanetitle, above] at (\xc,0|-LA.north) {C.~Independent calibration\\{\normalfont\color{rgGrayDk}$D_{\mathrm{cal}}$}};
\node[lanetitle, above] at (\xd,0|-LA.north) {D.~Locked decision\\{\normalfont\color{rgGrayDk}no new data}};
\node[lanetitle, above] at (\xe,0|-LA.north) {E.~Final evaluation\\{\normalfont\color{rgGrayDk}$D_{\mathrm{te}}$}};
\foreach \u/\v in {a1/a2,a2/a3,a3/a4,a4/a5,b1/b2,b2/b3,e1/e2,e2/e3,e3/e4,e4/e5} \draw[flow] (\u) -- (\v);
\draw[flow] (a5.east) -- ++(.16,0) |- (b1.west);
\draw[flow] (b3.east) -- ++(.14,0) coordinate (bj);
\foreach \c in {c1,c2,c3} \draw[flow] (bj) |- (\c.west);
\fill[rgGrayDk] (bj) circle (.9pt);
\coordinate (dj) at ($(c2.east)+(.2,0)$);
\foreach \c in {c1,c2,c3} \draw[draw=rgBlue, line width=.65pt] (\c.east) -| (dj);
\draw[bflow] (dj) -- (d0.west);
\fill[rgBlue] (dj) circle (1pt);
\draw[bflow] (d0.north) -- node[right, font=\scriptsize\bfseries, text=rgBlue]{yes} (dy.south);
\draw[flow, line width=.6pt] (d0.south) -- node[right, font=\scriptsize\bfseries, text=rgGrayDk]{no} (dn.north);
\coordinate (ej) at ($(e1.west)+(-.2,0)$);
\draw[flow] (dy.east) -| (ej);
\draw[flow] (dn.east) -| (ej);
\draw[flow] (ej) -- (e1.west);
\coordinate (T) at ($(LA.south)+(0,-.34)$);
\def\th{.15}
\fill[rgGrayL]    (LA.west|-T) rectangle ($(LA.east|-T)+(0,\th)$);
\fill[rgGrayL]    (LB.west|-T) rectangle ($(LB.east|-T)+(0,\th)$);
\fill[rgBlueL]    (LC.west|-T) rectangle ($(LC.east|-T)+(0,\th)$);
\draw[rgGray, line width=.4pt, pattern=north east lines, pattern color=rgGray]
                  (LE.west|-T) rectangle ($(LE.east|-T)+(0,\th)$);
\draw[rgInk, line width=.7pt] ($(LD.east|-T)+(.12,-.12)$) -- ($(LD.east|-T)+(.12,\th+.12)$);
\node[font=\scriptsize, text=rgGrayDk, align=center] at ($(LD.center|-T)+(0,\th/2)$) {no data accessed};
\node[font=\scriptsize, text=rgInk, anchor=south west, inner sep=1pt] at ($(LE.west|-T)+(0,\th+.02)$) {sealed until locked};
\draw[->, rgGray, line width=.4pt] ($(LA.west|-T)+(0,-.07)$) -- ($(LE.east|-T)+(.12,-.07)$);
\node[font=\scriptsize\bfseries, text=rgGrayDk, anchor=east] at ($(LA.west|-T)+(-.04,\th/2)$) {time};
\node[font=\scriptsize, text=rgGrayDk, below=1pt, align=center] at ($(LA.south|-T)+(0,-.07)$) {WDI 2001--2012 (2336 rows)};
\node[font=\scriptsize, text=rgGrayDk, below=1pt, align=center] at ($(LB.south|-T)+(0,-.07)$) {2013--2014 (398 rows)};
\node[font=\scriptsize, text=rgGrayDk, below=1pt, align=center] at ($(LC.south|-T)+(0,-.07)$) {2015--2016 (398 rows)\\blocks: 2015, 2016};
\node[font=\scriptsize, text=rgGrayDk, below=1pt, align=center] at ($(LE.south|-T)+(0,-.07)$) {2017--2021 (995 rows)};
\end{tikzpicture}}
\caption{\textbf{RAEGL deployment protocol.} Training fixes preprocessing, the global predictor $g$ and the candidate family. Selection fixes one candidate $h^\star$. Calibration evaluates three guards on $\Dcal$; the randomization guard also refits candidates on $\Dtr$/$\Dsel$ residuals with permuted labels (Fig.~\ref{fig:rand}). The action is locked before $\Dte$ is read, and evaluation quantities never feed back into phases A--D. Row counts refer to the WDI O3 study, whose lock used the calibration-maximum screen of \S\ref{sec:validity}.}
\label{fig:arch}
\end{figure}

\textbf{Data and candidates.} We observe $(x_{i,t},c_{i,t},y_{i,t})$ for units $i$ and years $t$, where $x\in\mathbb R^d$ are numerical predictors available at forecast time, $c$ is declared categorical context (e.g.\ Region, Income group) and $y$ is the target. The data are split chronologically,
\begin{equation}\label{eq:splits}
\Dtr\prec\Dsel\prec\Dcal\prec\Dte .
\end{equation}
A global predictor $g$ is fit on $\Dtr$ without context and frozen. Let $e_{i,t}=y_{i,t}-g(x_{i,t})$ be training residuals. For a context family $j\in\mathcal J$ with training categories $\mathcal C_j$, let $\phi_j(c)\in\{0,1\}^{K_j}$ be the one-hot code, $\mu_j$ its training mean, and $\tilde\phi_j(c)=\phi_j(c)-\mu_j$ for $c\in\mathcal C_j$ and $\tilde\phi_j(c)=0$ otherwise. For each penalty $\lambda\in\Lambda$,
\begin{equation}\label{eq:ridge}
\begin{aligned}(\hat b_{j\lambda},\hat\theta_{j\lambda})=\arg\min_{b,\theta}&\sum_{(i,t)\in\Dtr}\!\big[e_{i,t}-b-\tilde\phi_j(c_{i,t})^{\!\top}\theta\big]^2\\&+\lambda\|\theta\|_2^2,\end{aligned}
\end{equation}
with $b$ unpenalized. The candidate is $h_{j\lambda}(x,c)=g(x)+r_{j\lambda}(c)$ with $r_{j\lambda}(c)=\hat b_{j\lambda}+\tilde\phi_j(c)^\top\hat\theta_{j\lambda}$. Because the centered columns sum to zero over $\Dtr$, $\hat b_{j\lambda}=\overline e_{\mathrm{tr}}$ for every $(j,\lambda)$, so each correction is a common intercept shift plus a centered category deviation. An unseen category receives the intercept but no deviation; fallback is therefore a separate system-level action, $a=0$, that removes the whole correction. For a split $S$, the RMSE gain
\begin{equation}\label{eq:gain}
\Dh_S(h;g)=\mathrm{RMSE}_S(g)-\mathrm{RMSE}_S(h)
\end{equation}
is positive when the correction helps; \textsf{NoContext} denotes $h=g$.

\textbf{Selection.} With $M=|\mathcal J||\Lambda|$ declared pairs, selection uses only $\Dsel$:
\begin{equation}\label{eq:select}
(j^\star,\lambda^\star)=\arg\max_{(j,\lambda)}\Dh_{\mathrm{sel}}(h_{j\lambda};g),
\end{equation}
with \textsf{NoContext} chosen if no pair has positive gain. The selected $h^\star$ is then frozen.

\textbf{Deployment rule.} Let $\mathcal B$ be prespecified calibration blocks (in WDI, the years 2015 and 2016). RAEGL sets $a=1$ if and only if
\begin{equation}\label{eq:rule}
\begin{cases}\Dh_{\mathrm{sel}}(h^\star;g)>0, & p\le\alpha,\\[2pt] \Dh_{\mathrm{cal}}(h^\star;g)\ge\eprac, & \min_{\tau\in\mathcal B}\Dh_{\mathrm{cal},\tau}(h^\star;g)>0,\end{cases}
\end{equation}
and otherwise returns $\hat y=g(x)$ exactly. Here $p$ is the randomization $p$-value of \S\ref{sec:validity}, which tests a label-symmetry null and carries no statement about future harm. The floor and the block check are decision criteria without an inferential guarantee. The \emph{calibration threshold} is the rule that keeps only the first and third conditions.

\textbf{Outcomes.} Let $f_a=(1-a)g+ah^\star$, so $\Dh_{\mathrm{te}}(f_a;g)=a\,\Dh_{\mathrm{te}}(h^\star;g)$. A \emph{false deployment} (FD) is $a=1$ with $\Dh_{\mathrm{te}}(h^\star;g)<0$; a \emph{conservative miss} is $a=0$ with $\Dh_{\mathrm{te}}(h^\star;g)>0$. Regret against the better of the two policies is
\begin{equation}\label{eq:regret}
\mathrm{Regret}(f_a)=(1-a)\,[\Dh_{\mathrm{te}}]_+ + a\,[-\Dh_{\mathrm{te}}]_+ .
\end{equation}
The unconditional frequency $\Pr(a=1,\Dh_{\mathrm{te}}<0)$ and the conditional risk $\Pr(\Dh_{\mathrm{te}}<0\mid a=1)$ answer different questions: a rule can lower the former simply by deploying less.

\textbf{Squared-loss scale.} For an additive correction $r$, the per-row squared-loss improvement is $d=2er(c)-r(c)^2$. Since RMSE is monotone in MSE, $\Dh_S>0$ exactly when the mean of $d$ over $S$ is positive; the two scales agree in sign but not in magnitude. The country-cluster lower confidence bound (LCB) comparator of \S\ref{sec:setup} works on the mean of $d$, so it differs from the threshold through its margin, not its sign convention.

\textbf{Incremental value of a guard.} Consider tasks $i=1,\dots,n$ with realized test gains $d_i^{\mathrm{te}}=\Dh_{\mathrm{te},i}(h_i^\star;g_i)$, a threshold action $t_i$, and a stricter action $f_i\le t_i$ obtained by adding guards. Write $v_i=t_i-f_i\in\{0,1\}$ for the marginal veto, $G(a)=n^{-1}\sum_ia_id_i^{\mathrm{te}}$ for mean gain and $R(a)$ for mean regret.

\begin{proposition}[Nested-veto identity]\label{prop:increment}
$n\{G(f)-G(t)\}=-\sum_iv_id_i^{\mathrm{te}}=-n\{R(f)-R(t)\}$; the extra guards remove $\sum_iv_i\mathbf 1\{d_i^{\mathrm{te}}<0\}$ false deployments and add $\sum_iv_i\mathbf 1\{d_i^{\mathrm{te}}>0\}$ misses. With error weights $c_h,c_m$, the weighted loss changes by $n^{-1}\sum_iv_i(c_m[d_i^{\mathrm{te}}]_+-c_h[-d_i^{\mathrm{te}}]_+)$.
\end{proposition}
The identity is elementary (Appendix~\ref{app:proofs}). Its use is that it places the entire effect of an extra guard on the candidates it vetoes: with equal weights, the guard lowers expected regret only if $\mathbb E[v\,d^{\mathrm{te}}]<0$. A valid label test does not imply this inequality.

\section{The Permutation Screen under Panel Dependence}\label{sec:validity}
\textbf{The pipeline as a functional.} Hold fixed $g$, the residuals on $\Dtr\cup\Dsel\cup\Dcal$, the splits, the grid $\mathcal J\times\Lambda$ and the tie-breaking rule. The remaining input is the label vector $\mathbf c$ for all rows in $\Dtr\cup\Dsel\cup\Dcal$. Define
\begin{equation}\label{eq:functional}
T_{\mathrm{sel}}(\mathbf c)=\Dh_{\mathrm{cal}}\big(h_{\hat\jmath(\mathbf c),\hat\lambda(\mathbf c)};g\big),
\end{equation}
where $(\hat\jmath(\mathbf c),\hat\lambda(\mathbf c))$ is the output of \eqref{eq:select} when all candidates are refit with labels $\mathbf c$, and $T_{\mathrm{sel}}=0$ if \textsf{NoContext} is selected. Also let $T_{\max}(\mathbf c)=\max\{0,\max_{j,\lambda}\Dh_{\mathrm{cal}}(h_{j\lambda}(\mathbf c);g)\}$ and, for the observed winner, $T_{\mathrm{fix}}(\mathbf c)=\Dh_{\mathrm{cal}}(h_{j^\star\lambda^\star}(\mathbf c);g)$. Given a finite group $\mathcal G$ of label permutations and i.i.d.\ uniform draws $\pi_1,\dots,\pi_Q\in\mathcal G$, the matched $p$-value for a predeclared functional $T$ is
\begin{equation}\label{eq:pmatch}
\pmatch=\frac{1+\sum_{q=1}^{Q}\mathbf 1\{T(\pi_q\mathbf c)\ge T(\mathbf c)\}}{Q+1}.
\end{equation}
The WDI lock used a different construction, which compares the observed $T_{\mathrm{sel}}$ with a $T_{\max}$ null: $\plegacy=(1+\sum_q\mathbf 1\{T_{\max}(\pi_q\mathbf c)\ge T_{\mathrm{sel}}(\mathbf c)\})/(Q+1)$. We call it the legacy construction.

\begin{assumption}[Label symmetry]\label{ass:sym}
Under $H_0$, conditional on the residuals and all fixed pipeline inputs, $\pi\mathbf c\overset{d}{=}\mathbf c$ for every $\pi\in\mathcal G$.
\end{assumption}

\begin{proposition}[Matched validity]\label{prop:valid}
Under Assumption~\ref{ass:sym}, if $\pi_1,\dots,\pi_Q$ are i.i.d.\ uniform on $\mathcal G$ and independent of $\mathbf c$, then for any predeclared deterministic functional $T$ and $\alpha\in(0,1)$, $\Pr_{H_0}(\pmatch\le\alpha)\le\lfloor\alpha(Q+1)\rfloor/(Q+1)\le\alpha$.
\end{proposition}
This is the standard random-permutation result \citep{hemerik2018exact,phipson2010permutation}; the probability is joint over the data and the Monte Carlo draws. It covers $T_{\mathrm{sel}}$ and $T_{\max}$, which apply one fixed map to every label vector. It does not cover $T_{\mathrm{fix}}$, whose winner was chosen with the observed labels.

\begin{proposition}[Legacy construction is conservative]\label{prop:legacy}
For every label vector $T_{\max}(\mathbf c)\ge T_{\mathrm{sel}}(\mathbf c)$. Hence $\plegacy\ge\pmatch(T_{\mathrm{sel}})$ on the same draws, and under Assumption~\ref{ass:sym}, $\Pr_{H_0}(\plegacy\le\alpha)\le\alpha$.
\end{proposition}
The comparison holds on the same draws from the same group; it says nothing about a change of group.

\textbf{Why rerun selection.} If calibration used fresh units with labels independent of those seen in selection, the fixed-winner test would be valid conditional on the selected pair. In a panel the same countries carry the same labels in every split (in WDI, all 199 countries keep their Region over 2001--2016), so label symmetry given the residuals need not survive conditioning on the selection event. Rerunning selection inside each permutation restores a single fixed functional at a cost of $(Q+1)M$ ridge fits. It does not by itself make Assumption~\ref{ass:sym} true.

\textbf{Which group.} $\Gy$ permutes rows independently within each year. It preserves annual category counts but produces labelings that agree with the observed Region map for only about 19.5\% of rows. If residuals contain persistent country effects, the observed $\mathbf c$ is not exchangeable with such labelings even when context carries no signal, and the screen becomes anti-conservative. $\Gc$ permutes the country-to-label map once and applies it to all years, preserving persistence and group sizes. In a simulation with randomly assigned labels Assumption~\ref{ass:sym} holds for $\Gc$ by construction. Region in WDI is fixed by geography, so the assumption is a modeling choice that the data cannot verify.

\begin{proposition}[Test-only reversal is not identifiable]\label{prop:reversal}
Let $P$ and $P'$ induce the same distribution of $(\Dtr,\Dsel,\Dcal)$, and let $a=a(\Dtr,\Dsel,\Dcal,\xi)\in\{0,1\}$ for auxiliary randomness $\xi$ independent of the data. If $\Pr_P(\Dh_{\mathrm{te}}>0)\ge1-\delta$ and $\Pr_{P'}(\Dh_{\mathrm{te}}<0)\ge1-\delta'$, then $\Pr_P(\text{miss})+\Pr_{P'}(\text{FD})\ge1-\delta-\delta'$.
\end{proposition}
The proof is a two-point argument \citep{tsybakov2009introduction}. The stable and test-only-reversal regimes of our simulation form such a pair, which bounds what any calibration-based rule can achieve there.

\section{Experimental Setup}\label{sec:setup}
\textbf{WDI O3.} The target is the annual log-change in primary-energy intensity for 199 countries, with 50 numerical predictors from the World Development Indicators \citep{worldbank2024wdi}: lagged indicator values, lagged changes, missingness indicators and two calendar-year terms. Missing values are median-imputed with training statistics. The panel is a retrospective extract, not a sequence of real-time releases. Splits are 2001--2012 / 2013--2014 / 2015--2016 / 2017--2021 (2336 / 398 / 398 / 995 rows). The backbone is a Random Forest \citep{breiman2001random} (500 trees, depth 16, \texttt{max\_features}=0.5) chosen among 48 configurations by selection RMSE; it never sees $\Dcal$ or $\Dte$. Context families are Region, IncomeGroup and their additive combination, each with $\lambda\in\{0.01,0.1,1,10,100\}$, giving 15 pairs plus \textsf{NoContext}. Candidates are fit to in-sample residuals. Gate settings are $\alpha=0.05$, $\eprac=3\times10^{-4}$ (about 0.5\% of the backbone's calibration RMSE) and blocks $\{2015,2016\}$; these were fixed for this application and are not proposed as transferable defaults. The lock used $Q=1000$ draws from $\Gy$ and the legacy construction, implemented as $Z_0\ge Z_{(951)}$ among the sorted null values. That cutoff admits up to 50 exceedances, i.e.\ $p\le51/1001\approx0.051$, slightly above nominal $\alpha$; the observed 48 exceedances pass under either form. Test uncertainty uses a 5000-replicate country-cluster bootstrap \citep{efron1979bootstrap,cameron2008bootstrap}.

\textbf{Timing of analyses.} The WDI O3 action was locked before 2017--2021 were read. The analyses below were run after those years had been opened and are reported as retrospective; none is used to revise the WDI decision. They are (a)~a sensitivity study that recomputes all 398 calibration predictions from the frozen backbone (maximum absolute difference $1.1\times10^{-16}$) and evaluates $T_{\mathrm{sel}}$, $T_{\max}$ and $T_{\mathrm{fix}}$ on $Q=999$ draws from each of $\Gc$ and $\Gy$; (b)~a residual and rolling-origin study; and (c)~the synthetic trials.

\textbf{Residual construction and rolling origins.} The forest has training RMSE 0.0328 but selection RMSE 0.0692, so in-sample residuals understate its error. Holding $g$, the grid, the floor, the blocks and $\alpha$ fixed, we refit every candidate on in-sample, out-of-bag (OOB; reconstructed exactly from the stored bootstrap seeds) and forward cross-fitted residuals $e_{i,t}=y_{i,t}-g_{<t}(x_{i,t})$, where $g_{<t}$ is the same forest specification refit on training years before $t$. We use the four origins O1--O4 of the protocol file (training ends 2008/2010/2012/2014; two selection and two calibration years each). O3 uses the frozen $g$; O1, O2 and O4 refit the O3 specification without retuning. The screen is matched $T_{\mathrm{sel}}$ under $\Gc$ with $Q=999$.

\textbf{Synthetic trials.} We simulate 96 countries over 16 years, split as years 0--6 / 7--9 / 10--12 / 13--15. The design crosses context strength $s\in\{0,0.45,1.10\}$, four drift mechanisms (stable, moderate decay, calibration decay, test-only reversal), three Region-imbalance settings and 20 seeds, for 720 trials. Outcomes include persistent country effects and AR(1) noise, and the true effect acts through Region only (Appendix~\ref{app:synthetic}). The backbone is a context-free Ridge regression; candidates are three families (Region, IncomeGroup, Region$\times$Income cells) times $\lambda\in\{0.5,10\}$. Settings are $Q=99$ draws from $\Gc$, $\alpha=0.05$, $\eprac=0.03$ and three yearly calibration blocks. We compare always-selected, the calibration threshold ($\Dh_{\mathrm{cal}}\ge0.03$), a one-sided 95\% country-cluster LCB on the mean calibration $d$, each guard alone, two-guard conjunctions and the full rule. All actions are fixed before test years are read. Trials are independent but not identically distributed across the 36 design cells; the normal intervals in Table~\ref{tab:policies} treat the design as fixed.

\section{Results}\label{sec:results}

\subsection{Validity depends on the permutation group}
\begin{figure}[t]
\centering
\includegraphics[width=\textwidth]{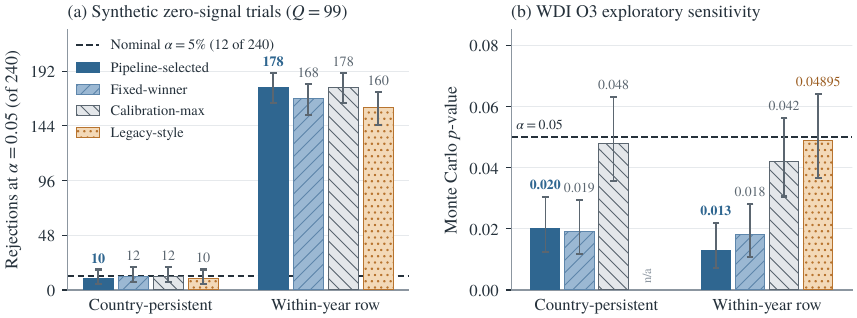}
\caption{\textbf{Randomization diagnostics.} Legacy-style compares a $T_{\max}$ null with the observed $T_{\mathrm{sel}}$; calibration-max uses $T_{\max}$ for both. (a) Synthetic zero-signal rejections on identical draws (Wilson 95\% CI). (b) WDI O3 $p$-values (Clopper--Pearson CI); the lock value uses $Q=1000$ and $\Gy$, the others $Q=999$.}
\label{fig:validity}
\end{figure}

In the 240 zero-signal synthetic trials, where Assumption~\ref{ass:sym} holds for $\Gc$ by design, the pipeline statistic rejects in 10 trials (4.2\%, Wilson 95\% CI [2.3, 7.5]\%; \citealp{wilson1927probable}), against 12 for fixed-winner and calibration-max and 10 for legacy-style (Fig.~\ref{fig:validity}a). Under $\Gy$ the pipeline statistic rejects in 178 trials (74.2\%, [68.3, 79.3]\%) and the other statistics in 160--178. Persistent country effects produce nonzero Region means without any context signal; $\Gc$ reproduces this aliasing in the null distribution and $\Gy$ removes it.

Within $\Gc$, the choice of statistic matters little. Re-selection and fixed-winner reject 10 and 12 null trials, and their power is 172 versus 173 at $s=0.45$ and 180 versus 180 at $s=1.10$. Across all 720 trials the matched decision differs from fixed-winner in 3 cases and from calibration-max in 6, and $\plegacy\ge\pmatch$ holds for all 711 contextual winners without changing any decision (Appendix Table~\ref{tab:ablation}). On WDI O3 all seven constructions give $p<0.05$ (pipeline: 0.020 under $\Gc$, 0.013 under $\Gy$; Appendix Table~\ref{tab:constructions}). Because Region is geographic, these values are descriptive: they do not establish Type-I control on the observational panel.

\subsection{What the additional guards add}
\begin{table}[t]
\centering\small
\caption{\textbf{Policies over 720 synthetic trials.} FD: harmful deployment, per trial and per deployment. Miss is per trial; mean gain and regret average over all trials (normal 95\% CI). Deployment rates differ across rows. Bold: best value per column (lowest harm, misses and regret; highest gain); deployment rate has no preferred direction. Per-drift FD counts are in Appendix Table~\ref{tab:fddrift}.}
\label{tab:policies}
\setlength{\tabcolsep}{4pt}
\begin{tabular}{@{}lrrrrlr@{}}
\toprule
Policy & Deploy & FD all & FD / deployed & Miss & Mean gain [95\% CI] & Regret\\
\midrule
Always selected & 98.8\% & 42.1\% & 42.6\% & \textbf{0.0\%} & $-0.0617$ [$-0.082$, $-0.041$] & 0.1197\\
Calibration threshold & 49.4\% & 18.5\% & 37.4\% & 25.7\% & $-0.0241$ [$-0.043$, $-0.005$] & 0.0821\\
Country-cluster LCB & 51.8\% & \textbf{17.9\%} & 34.6\% & 22.8\% & $-0.0233$ [$-0.043$, $-0.004$] & 0.0813\\
Randomization only & 50.3\% & 18.5\% & 36.7\% & 24.9\% & $-0.0240$ [$-0.043$, $-0.005$] & 0.0820\\
Temporal only & 68.3\% & 21.0\% & \textbf{30.7\%} & 9.3\% & $\mathbf{-0.0224}$ [$-0.042$, $-0.003$] & \textbf{0.0804}\\
Full RAEGL & 49.2\% & 18.5\% & 37.6\% & 26.0\% & $-0.0242$ [$-0.043$, $-0.005$] & 0.0822\\
\bottomrule
\end{tabular}
\end{table}

Every gating rule lowers harmful deployment mainly by deploying less (Table~\ref{tab:policies}). The full rule and the calibration threshold deploy 354 and 356 times and produce the same 133 harmful deployments. Because the actions are nested, Proposition~\ref{prop:increment} attributes their whole difference to two vetoes, both of corrections with positive test gain (0.02229 and 0.01686). The full rule therefore prevents no additional harm, adds two misses and raises mean regret by about $5.44\times10^{-5}$. After the floor and the screen, the block check vetoes nothing. Of the 357 rejected contextual winners, 355 fail the floor and 349 fail the screen.

The other rules sit at different operating points. The LCB deploys in 51.8\% of trials with 17.9\% unconditional harm, and its paired mean gain exceeds that of the full rule by $9.4\times10^{-4}$ (SE $3.7\times10^{-4}$). Temporal-only deploys 492 times, with 151 harmful deployments (full rule: 133), 67 misses (187) and lower regret (0.0804 versus 0.0822). No rule is better on harm, misses and coverage at once, and a comparison at equal coverage needs paired trial-level outcomes that we do not have (\S\ref{sec:discussion}). A post hoc sweep of $(\alpha,\eprac)$ is given in Appendix Table~\ref{tab:thresh}.

\subsection{WDI O3: the locked decision}
\begin{table}[t]
\centering\small
\caption{\textbf{WDI O3 locked decision.} Gains are RMSE differences ($\times10^{-4}$), recomputed from the frozen prediction files. Test CI: 5000-replicate country-cluster bootstrap. $^{a}$The lock cutoff corresponds to $p\le0.051$ (\S\ref{sec:setup}).}
\label{tab:wdi}
\setlength{\tabcolsep}{6pt}
\begin{tabular}{@{}lrrl@{}}
\toprule
Component & Observed & Required & Status\\
\midrule
Selection gain $\Dh_{\mathrm{sel}}$ & 1.6314 & $>0$ & pass\\
Calibration gain $\Dh_{\mathrm{cal}}$ & 1.1903 & $>0$ & pass\\
Randomization, $\plegacy$ & 0.04895 & $\le0.05^{a}$ & narrow pass\\
Practical floor & 1.1903 & $\ge3.0000$ & \textbf{fail}\\
Worst block (2016) & $-1.0445$ & $>0$ & \textbf{fail}\\
\midrule
Action & \multicolumn{3}{l}{exact fallback to $g$}\\
Test gain of rejected $h^\star$ & \multicolumn{3}{l}{0.9222; CI $[-0.091, 2.028]$}\\
\bottomrule
\end{tabular}
\end{table}

Selection picks Region with $\lambda=10$ (selection RMSE $0.0692045\to0.0690414$); the other Region penalties follow within $10^{-6}$. The candidate gains $1.1903\times10^{-4}$ on calibration. Under the legacy screen 48 of 1000 null values reach this gain, so $\plegacy=49/1001=0.04895$, with a Monte Carlo 95\% interval of $[0.036,0.063]$ for the exceedance probability. The screen result is thus within Monte Carlo error of the boundary, but the decision does not depend on it: the floor and the 2016 block both fail, and RAEGL returns the forest exactly (Table~\ref{tab:wdi}).

On 2017--2021 the global and rejected-candidate RMSEs are 0.055576 and 0.055483, so fallback forgoes $9.22\times10^{-5}$, or 0.17\% of test RMSE. The country-cluster interval crosses zero, and 0.02\% of bootstrap replicates reach the floor. Year-level gains are $-1.00$, $0.00$, $3.48$, $1.61$ and $0.49$ ($\times10^{-4}$), so the first test year is harmful, as was the second calibration year. 105 countries improve and 94 worsen; only Sub-Saharan Africa (positive) and East Asia \& Pacific (negative) have regional intervals excluding zero (Appendix Table~\ref{tab:regions}). The evidence points to a small, heterogeneous effect below the declared practical standard.

The backbone itself is weak out of sample. A zero-change forecast has test RMSE 0.055508 and the training-mean forecast has calibration RMSE 0.055557, both no worse than $g$ (Appendix Table~\ref{tab:naive}). RAEGL decides whether to add a component to a declared $g$; whether $g$ should be deployed is a separate question, and WDI O3 is a low-signal setting for it.

\subsection{Residual construction and rolling origins}\label{sec:residual}
\begin{table}[t]
\centering
\caption{\textbf{WDI origin--residual cells (retrospective).} Matched $T_{\mathrm{sel}}$, $\Gc$, $Q=999$; floor $3\times10^{-4}$; gains $\times10^{-4}$. In/OOB/Fwd: in-sample, out-of-bag, forward residuals; R/RI: Region, Region+Income; D/F: deploy/fallback. Brackets: 5000-replicate country-cluster 95\% CI per cell. The calibration threshold gives the same action in all 12 cells.}
\label{tab:residual}
\setlength{\tabcolsep}{5pt}\small
\begin{tabular}{@{}lllrrrcr@{}}
\toprule
 & Res. & Selected & $\Dh_{\mathrm{cal}}$ & min blk & $p$ & Act. & Test gain [95\% CI]\\
\midrule
O1 & In & RI, .01 & $-0.44$ & $-1.15$ & .799 & F & $1.04$ [$-0.47$, $2.49$]\\
 & OOB & RI, .01 & $-1.73$ & $-2.17$ & .874 & F & $0.20$ [$-3.03$, $3.28$]\\
 & Fwd & \textsf{NoContext} & -- & -- & -- & F & 0\\
O2 & In & R, 100 & $2.08$ & $1.38$ & .009 & F & $0.83$ [$-0.12$, $1.81$]\\
 & OOB & R, 100 & $3.26$ & $2.38$ & .010 & \textbf{D} & $0.94$ [$-0.81$, $2.70$]\\
 & Fwd & R, 10 & $4.10$ & $3.05$ & .006 & \textbf{D} & $-0.16$ [$-2.75$, $2.41$]\\
O3 & In & R, 10 & $1.19$ & $-1.04$ & .019 & F & $0.92$ [$-0.11$, $2.01$]\\
 & OOB & R, 100 & $1.53$ & $-2.48$ & .022 & F & $0.44$ [$-1.22$, $2.19$]\\
 & Fwd & RI, 100 & $-1.47$ & $-8.87$ & .480 & F & $-3.83$ [$-6.92$, $-0.77$]\\
O4 & In & RI, .01 & $-1.50$ & $-1.57$ & .951 & F & $1.77$ [$0.26$, $3.30$]\\
 & OOB & R, 100 & $-1.17$ & $-1.34$ & .822 & F & $1.61$ [$-0.15$, $3.40$]\\
 & Fwd & \textsf{NoContext} & -- & -- & -- & F & 0\\
\bottomrule
\end{tabular}
\end{table}

In-sample fitting distorts the residual scale: in-sample residual RMSE is 0.030--0.035, against 0.063--0.067 for OOB and forward residuals, close to the selection RMSE. The O3 action is the same under all three constructions. The OOB candidate again fails the floor and the 2016 block; the forward candidate has negative calibration gain and would have lost $3.83\times10^{-4}$ (0.69\% of test RMSE, interval excluding zero), and every rule rejects it.

The gate deploys in 2 of the 12 cells (O2 with OOB and forward residuals), with test gains of $+0.94$ and $-0.16\times10^{-4}$, neither distinguishable from zero. The selected correction has positive test gain in 8 of the 10 cells with a contextual winner, but only O4 in-sample has an interval excluding zero, and its calibration gain is negative ($-1.50$). This is a beneficial calibration-to-test reversal, the mirror image of the case in Proposition~\ref{prop:reversal}. Separating the shared intercept $\hat b=\overline e_{\mathrm{tr}}$ from the category deviation shows that the intercept lowers test gain in 9 of 12 cells (Appendix Table~\ref{tab:residextra}); we report this but do not change the rule after seeing test data. In a zero-floor comparison on the same cells, positive-gain-only and randomization-only rules each deploy in 5 cells and the floor-only and full rules in 2, with one harmful deployment in every case (Appendix Table~\ref{tab:wdiablate}). The overlapping windows make these cells dependent, so they illustrate rather than test the rules.

\subsection{Drift and test-only reversal}
The gated rules reject nearly every candidate when the effect decays before calibration (full rule: 1 harmful deployment in 180 trials, always-selected: 133). They cannot prevent the 120 harmful deployments under strong-signal test-only reversal, which is the empirical counterpart of Proposition~\ref{prop:reversal}. This regime receives 25\% of the synthetic trials by design, so the pooled harm rates in Table~\ref{tab:policies} depend on that weight and do not estimate a real-world frequency (Appendix~\ref{app:synthetic}).

\section{Discussion and Limitations}\label{sec:discussion}
\textbf{What reduces harm.} In our simulation the full rule lowers unconditional harmful deployment from 42.1\% to 18.5\%, and the calibration threshold alone achieves the same reduction. The useful ingredient is an independent calibration period with a declared minimum gain. By Proposition~\ref{prop:increment}, an additional guard pays off only if the candidates it vetoes tend to have negative future gain; in 718 of 720 trials and all 12 WDI cells it vetoed nothing.

\textbf{What the screen tests.} The permutation screen tests a label-symmetry null. Rejecting it, clearing a practical floor and improving future risk are three different claims. Country-persistent permutation is justified in the simulation, where labels are randomized; for geographic labels it is an assumption, and within-year shuffling should not be used when units carry persistent effects.

\textbf{Limitations.} The WDI backbone does not beat naive forecasts, and the rolling origins share years and countries, so the real-panel results describe one low-signal application rather than validate the rule prospectively. The 25\% weight of the reversal regime is a stress-test choice. Repeated reapproval over successive horizons has no error control here. Finally, the per-trial decisions and the generating script of the synthetic study were not retained. Appendix~\ref{app:synthetic} specifies the full generative model, and Tables~\ref{tab:policies}, \ref{tab:ablation} and \ref{tab:allpolicies} are computed from stored aggregate counts; extending the comparison, for example to equal-coverage risk curves, requires a fresh paired run.

\section{Conclusion}\label{sec:conclusion}
Deciding whether to deploy a contextual correction is distinct from selecting it. In our experiments the decision is driven by independent calibration: a plain threshold on calibration gain reproduces the harm reduction of the full rule, and the additional guards vetoed only beneficial corrections. A permutation screen is valid only for a group that respects the dependence structure of the panel, and no rule that sees only pre-deployment data can protect against a reversal that begins at deployment. Additional guards should be justified by evidence that they veto harmful corrections, not by the significance of a label test.

\bibliographystyle{plainnat}
\bibliography{references}

\clearpage
\appendix
\setlength{\intextsep}{8pt plus 2pt minus 2pt}
\setlength{\textfloatsep}{10pt plus 2pt minus 2pt}
\setlength{\floatsep}{8pt plus 2pt minus 2pt}

\section{Algorithm}\label{app:algo}
\begin{algorithm}[H]
\caption{RAEGL (one deployment horizon)}
\begin{algorithmic}[1]
\Require chronological splits; family $\mathcal J\times\Lambda$; $\alpha,\eprac,\mathcal B$; group $\mathcal G$; $Q$
\State Fit preprocessing and $g$ on $\Dtr$ (no context); compute residuals $e$ on $\Dtr\cup\Dsel\cup\Dcal$
\State \textbf{function} $T(\mathbf c)$: fit all $r_{j\lambda}$ on $(\Dtr,\mathbf c)$ via \eqref{eq:ridge}; select via \eqref{eq:select} on $\Dsel$; \Return $\Dh_{\mathrm{cal}}$ of the selected candidate (0 if \textsf{NoContext})
\State $h^\star\gets$ selected candidate under observed $\mathbf c$; $Z_0\gets T(\mathbf c)$
\For{$q=1,\dots,Q$} draw $\pi_q\sim\mathrm{Unif}(\mathcal G)$; $Z_q\gets T(\pi_q\mathbf c)$ \EndFor
\State $p\gets(1+\#\{q:Z_q\ge Z_0\})/(Q+1)$
\State $a\gets\mathbf 1\{\Dh_{\mathrm{sel}}(h^\star)>0\}\cdot\mathbf 1\{p\le\alpha\}\cdot\mathbf 1\{\Dh_{\mathrm{cal}}(h^\star)\ge\eprac\}\cdot\mathbf 1\{\min_{\tau\in\mathcal B}\Dh_{\mathrm{cal},\tau}(h^\star)>0\}$
\State \textbf{lock} $a$ (record hash); deploy $f_a=(1-a)g+ah^\star$; only then open $\Dte$ and report \eqref{eq:regret}
\end{algorithmic}
\end{algorithm}
The screen costs $(Q+1)M$ ridge fits of $K_j$-dimensional categorical designs plus scoring; memory is one replicate plus $Q$ scalars.

\begin{figure}[!htbp]
\centering
\resizebox{0.9\textwidth}{!}{
\begin{tikzpicture}[
  font=\scriptsize, >={Stealth[length=3.6pt,width=3.2pt]}, text=rgInk,
  st/.style={draw=rgGray, line width=.4pt, rounded corners=1.5pt, fill=white, align=center,
             minimum height=.78cm, text width=2.0cm, inner sep=2pt},
  obs/.style={st, draw=rgBlue, line width=.6pt},
  rnd/.style={st, draw=rgGray, line width=.45pt},
  stat/.style={circle, line width=.6pt, fill=white, minimum size=.78cm, inner sep=0pt, font=\small},
  flow/.style={->, draw=rgBlue, line width=.55pt},
  gflow/.style={->, draw=rgGrayDk, line width=.5pt},
  note/.style={rounded corners=2pt, align=left, text width=4.2cm, inner sep=3pt, font=\scriptsize, line width=.4pt},
]
\def\dx{2.45}
\node[obs] (o0) at (0,0) {Observed labels\\$\mathbf c=(c_i)_{i}$};
\node[obs] (o1) at (\dx,0) {Refit all $r_{j,\lambda}$\\on $D_{\mathrm{tr}}$};
\node[obs] (o2) at (2*\dx,0) {Select $(j^\star\!,\lambda^\star)$\\on $D_{\mathrm{sel}}$};
\node[obs] (o3) at (3*\dx,0) {Score \emph{selected}\\candidate on $D_{\mathrm{cal}}$};
\node[stat, draw=rgBlue, fill=rgBluePale] (z0) at (4*\dx-.3,0) {$Z_0$};
\node[rnd] (r0) at (0,-1.55) {Permuted labels\\$\mathbf c^{(q)}=\pi_q\mathbf c$};
\node[rnd] (r1) at (\dx,-1.55) {Refit all $r_{j,\lambda}$\\on $D_{\mathrm{tr}}$};
\node[rnd] (r2) at (2*\dx,-1.55) {Re-select $(j^{(q)}\!,\lambda^{(q)})$\\on $D_{\mathrm{sel}}$};
\node[rnd] (r3) at (3*\dx,-1.55) {Score \emph{re-selected}\\candidate on $D_{\mathrm{cal}}$};
\node[stat, draw=rgGrayDk] (zq) at (4*\dx-.3,-1.55) {$Z_q$};
\foreach \u/\v in {o0/o1,o1/o2,o2/o3,o3/z0} \draw[flow] (\u) -- (\v);
\foreach \u/\v in {r0/r1,r1/r2,r2/r3,r3/zq} \draw[gflow] (\u) -- (\v);
\foreach \i in {1,2,3} \draw[rgGrayL, line width=.5pt, dash pattern=on 1.2pt off 1.2pt] (o\i.south) -- (r\i.north);
\draw[gflow] (o0) -- node[left, align=right, font=\tiny, text=rgGrayDk]{$\pi_q\sim\mathrm{Unif}(\mathcal G)$\\i.i.d., $q=1,\dots,Q$} (r0);
\begin{scope}[on background layer]
\node[draw=rgGray, fill=rgGrayPale, dashed, rounded corners=3pt, fit=(r0)(zq), inner sep=4pt,
      label={[font=\tiny, text=rgGrayDk, anchor=north east]south east:repeat $Q$ times}] (plate) {};
\end{scope}
\node[font=\tiny, text=rgGrayDk, align=center, above=1pt of o2] {same $g$, residuals, splits, grid $\mathcal J{\times}\Lambda$, tie rule};
\node[draw=rgBlue, line width=.75pt, rounded corners=2pt, fill=rgBluePale, align=center, text width=3.25cm, inner sep=3pt]
  (pv) at (4*\dx+2.25,-.78) {$p_{\mathrm{match}}=\dfrac{1+\sum_{q=1}^{Q}\mathbf 1\{Z_q\ge Z_0\}}{Q+1}$\\[2pt]
  {\tiny $Z_0$ and every $Z_q$ are the \emph{same} functional $T(\cdot)$ of the label vector}};
\draw[flow] (z0.east) -- ++(.25,0) |- ($(pv.west)+(0,.12)$);
\draw[gflow] (zq.east) -- ++(.25,0) |- ($(pv.west)+(0,-.12)$);
\node[note, draw=rgBlue, fill=rgBluePale, anchor=north west] (n1) at (-1.05,-2.62) {\textbf{\textcolor{rgBlue}{$\mathcal G_{\mathrm{country}}$} (persistent):} one permutation of the country$\to$label map, applied to every year. Preserves label persistence and group sizes. Valid if $\mathbf c\overset{d}{=}\pi\mathbf c$ for $\pi\in\mathcal G_{\mathrm{country}}$ under $H_0$; for observational WDI this is an \emph{assumption}.};
\node[note, draw=rgGray, fill=rgGrayPale, anchor=north west] (n2) at ($(n1.north east)+(.2,0)$) {\textbf{$\mathcal G_{\mathrm{year}}$ (within-year rows):} an independent permutation of rows in every year. Preserves annual counts but destroys persistence (WDI: 100\% of countries keep their Region; only $\approx$19.5\% agreement expected after permutation). Diagnostic only.};
\node[note, draw=rgOrange, fill=rgOrangePale, anchor=north west] (n3) at ($(n2.north east)+(.2,0)$) {\textbf{\textcolor{rgOrangeDk}{Legacy (WDI lock):}} $Z_0^{\mathrm{sel}}$ compared with $Z_q^{\max}=\max\{0,\max_{j,\lambda}\widehat\Delta^{(q)}_{\mathrm{cal}}\}$ under $\mathcal G_{\mathrm{year}}$; no re-selection on $D_{\mathrm{sel}}$. Since $Z_q^{\max}\ge Z_q^{\mathrm{sel}}$, it is a conservative, \emph{different} construction (Prop.~\ref{prop:legacy}).};
\end{tikzpicture}}
\caption{\textbf{Matched randomization.} The observed and permuted pipelines apply the same functional $T$: refit all residual candidates on $\Dtr$, re-select on $\Dsel$, and score the selected candidate on $\Dcal$. The global predictor and residuals are held fixed because $g$ does not use context. The two groups encode different null hypotheses. The legacy statistic used for the WDI lock is shown bottom right.}
\label{fig:rand}
\end{figure}

\section{Proofs}\label{app:proofs}
\textbf{Proof of Proposition~\ref{prop:increment}.} By construction $f_i=t_i-v_i$. Substituting into $G$ gives the gain identity; using $[d]_+-[-d]_+=d$ in the regret gives the regret identity, and subtracting the indicator sums gives the FD and miss counts. The weighted loss follows by the same substitution. $\square$

For the remaining results, $U$ denotes every fixed pipeline input (the frozen $g$, the residuals on $\Dtr\cup\Dsel\cup\Dcal$, the splits, the grid and the tie-breaking rule), and all probabilities are conditional on $U$. An admissible statistic $T$ is a single predeclared deterministic map from label vectors to $\mathbb R$ for fixed $U$; this covers $T_{\mathrm{sel}}$ and $T_{\max}$ but not $T_{\mathrm{fix}}$. The draws $\pi_1,\dots,\pi_Q$ are i.i.d.\ uniform on the finite group $\mathcal G$ and independent of $\mathbf c$. Write $Z_0=T(\mathbf c)$, $Z_q=T(\pi_q\mathbf c)$ and $R_i=\#\{q\in\{0,\dots,Q\}:Z_q\ge Z_i\}$, so $\pmatch=R_0/(Q+1)$.

\textbf{Lemma 1 (counting).} For every $(Z_0,\dots,Z_Q)$ and integer $k\ge0$, $\#\{i:R_i\le k\}\le k$.

\emph{Proof.} Let $S=\{i:R_i\le k\}$ and suppose $|S|>k$. Choose $i^\ast\in S$ with $Z_{i^\ast}=\min_{i\in S}Z_i$. Every $j\in S$ has $Z_j\ge Z_{i^\ast}$, so $R_{i^\ast}\ge|S|>k$, a contradiction. Ties are handled because $R_i$ counts all $q$ with $Z_q\ge Z_i$. $\square$

\textbf{Lemma 2 (exchangeability).} Under Assumption~\ref{ass:sym}, $(Z_0,\dots,Z_Q)$ is exchangeable.

\emph{Proof.} Let $\sigma\sim\mathrm{Unif}(\mathcal G)$ be independent of $(\mathbf c,\pi_1,\dots,\pi_Q)$. For fixed $s\in\mathcal G$, Assumption~\ref{ass:sym} and independence of $(\pi_q)$ give $(s\mathbf c,\pi_{1:Q})\overset{d}{=}(\mathbf c,\pi_{1:Q})$; averaging over $s$ gives $(\sigma\mathbf c,\pi_{1:Q})\overset{d}{=}(\mathbf c,\pi_{1:Q})$. Applying $(\mathbf v,\pi_{1:Q})\mapsto(T(\mathbf v),T(\pi_1\mathbf v),\dots,T(\pi_Q\mathbf v))$,
\[(Z_0,\dots,Z_Q)\overset{d}{=}\big(T(\rho_0\mathbf c),\dots,T(\rho_Q\mathbf c)\big),\qquad\rho_0=\sigma,\ \rho_q=\pi_q\sigma .\]
Conditional on $\mathbf c$, $(\rho_0,\dots,\rho_Q)$ is i.i.d.\ uniform on $\mathcal G$, since $\Pr(\rho_0=g_0,\rho_q=g_q\ \forall q)=\Pr(\sigma=g_0)\prod_q\Pr(\pi_q=g_qg_0^{-1})=|\mathcal G|^{-(Q+1)}$. The right-hand side is therefore exchangeable \citep[cf.][Theorem~2]{hemerik2018exact}. $\square$

\textbf{Proof of Proposition~\ref{prop:valid}.} By Lemma~2, $\Pr(R_0\le k)=\Pr(R_i\le k)$ for every $i$, so by Lemma~1
\[\Pr(R_0\le k)=\frac{1}{Q+1}\,\mathbb E\,\#\{i:R_i\le k\}\le\frac{k}{Q+1}.\]
Since $\pmatch\le\alpha$ iff $R_0\le\lfloor\alpha(Q+1)\rfloor$, the claim follows. Deterministic tie-breaking inside $T$ is part of $U$. $\square$

\textbf{Proof of Proposition~\ref{prop:legacy}.} $T_{\max}\ge0$. If \textsf{NoContext} is selected under $\mathbf v$, $T_{\mathrm{sel}}(\mathbf v)=0\le T_{\max}(\mathbf v)$; otherwise the selected pair is one of the maximized pairs. Applying the inequality at $\mathbf v=\pi_q\mathbf c$ for each draw gives $\mathbf 1\{T_{\max}(\pi_q\mathbf c)\ge T_{\mathrm{sel}}(\mathbf c)\}\ge\mathbf 1\{T_{\mathrm{sel}}(\pi_q\mathbf c)\ge T_{\mathrm{sel}}(\mathbf c)\}$, hence $\plegacy\ge\pmatch(T_{\mathrm{sel}})$. Validity follows from Proposition~\ref{prop:valid} applied to $T_{\mathrm{sel}}$. $\square$

\textbf{Proof of Proposition~\ref{prop:reversal}.} Since $(\Dtr,\Dsel,\Dcal)$ and $\xi$ have the same law under $P$ and $P'$, so does $a$; let $\omega=\Pr_P(a=1)=\Pr_{P'}(a=1)$. Using $\Pr(A\cap B)\ge\Pr(A)-\Pr(B^c)$,
\[\Pr\nolimits_P(\text{miss})\ge(1-\omega)-\delta,\qquad \Pr\nolimits_{P'}(\text{FD})\ge\omega-\delta'.\]
Adding the two bounds eliminates $\omega$. $\square$

\section{WDI O3 Details}\label{app:wdi}
\textbf{Country-level tests.} For country $i$ let $d_i=\sum_{t:(i,t)\in\Dte}[(y_{it}-g(x_{it}))^2-(y_{it}-h^\star(x_{it},c_i))^2]$, with $\sum_id_i=0.0101904$. A Monte Carlo sign-flip test of the weighted squared-error gains ($B=10{,}000$, add-one correction) gives one-sided $p=0.039$ and two-sided $p=0.083$; it assumes a sign symmetry that observational data do not establish. The exact binomial sign test on 105 better and 94 worse countries gives $p=0.239$ (one-sided) and $0.478$ (two-sided). The two tests answer different questions, and neither bears on the locked action.

\begin{table}[H]
\centering\small
\caption{Regional test diagnostics for the rejected WDI O3 candidate (exploratory, no multiplicity correction). Gains and bounds $\times10^{-4}$. Intervals use within-region country-cluster resampling (3000 replicates). North America has 3 countries, so its interval is unstable. LORO: overall test gain when the region is left out.}
\label{tab:regions}
\setlength{\tabcolsep}{4pt}\begin{tabular}{@{}lrrrrr@{}}
\toprule
Region & Cal. & Test & 95\% CI & Better/worse & LORO\\
\midrule
Sub-Saharan Africa & 3.756 & 3.358 & [1.040, 5.287] & 30/17 & 0.291\\
Europe \& Central Asia & $-0.557$ & 1.810 & [$-2.405$, 6.428] & 27/22 & 0.764\\
Middle East \& N.\ Africa & $-0.733$ & 1.164 & [$-0.654$, 3.262] & 13/9 & 0.888\\
Latin America \& Caribbean & 2.550 & 0.913 & [$-0.861$, 2.939] & 21/17 & 0.927\\
North America & 0.515 & $-0.134$ & [$-5.200$, 3.999] & 1/2 & 0.936\\
South Asia & $-2.647$ & $-1.433$ & [$-3.604$, 3.371] & 3/3 & 0.999\\
East Asia \& Pacific & 0.149 & $-1.657$ & [$-2.283$, $-1.085$] & 10/24 & 1.664\\
\bottomrule
\end{tabular}
\end{table}

\textbf{Randomization constructions.} In the legacy null, \textsf{NoContext} attains the maximum in 483 of 1000 replicates. Null quantiles are: median $1.8\times10^{-6}$, 90\% $8.8\times10^{-5}$, 95\% $1.1694\times10^{-4}$, 99\% $1.970\times10^{-4}$, maximum $3.666\times10^{-4}$. Under $\Gc$ and $\Gy$ the re-selected null picks \textsf{NoContext} in 18.0\% and 27.8\% of replicates. Calibration block gains are $+2.932\times10^{-4}$ (2015) and $-1.044\times10^{-4}$ (2016).

\begin{table}[H]
\centering\small
\caption{Randomization constructions on WDI O3. $Z_0^{\mathrm{sel}}=1.190293\times10^{-4}$; $Z_0^{\max}=1.200827\times10^{-4}$. Exceedance CIs are Clopper--Pearson 95\% \citep{clopper1934use}. The first row is the locked screen; the others were computed after the test years were opened. Fixed-winner rows are not covered by Proposition~\ref{prop:valid}.}
\label{tab:constructions}
\setlength{\tabcolsep}{4pt}\begin{tabular}{@{}llllrrl@{}}
\toprule
Construction & Group & Observed & Null & $Q$ & Exceed. & $p$ [CI of exceedance prob.]\\
\midrule
Legacy (lock) & $\Gy$ & $Z_0^{\mathrm{sel}}$ & $T_{\max}$ & 1000 & 48 & 0.04895 [0.036, 0.063]\\
Pipeline-selected & $\Gc$ & $Z_0^{\mathrm{sel}}$ & $T_{\mathrm{sel}}$ & 999 & 19 & 0.020 [0.011, 0.030]\\
Calibration-max & $\Gc$ & $Z_0^{\max}$ & $T_{\max}$ & 999 & 47 & 0.048 [0.035, 0.062]\\
Fixed-winner & $\Gc$ & $Z_0^{\mathrm{sel}}$ & $T_{\mathrm{fix}}$ & 999 & 18 & 0.019 [0.011, 0.028]\\
Pipeline-selected & $\Gy$ & $Z_0^{\mathrm{sel}}$ & $T_{\mathrm{sel}}$ & 999 & 12 & 0.013 [0.006, 0.021]\\
Calibration-max & $\Gy$ & $Z_0^{\max}$ & $T_{\max}$ & 999 & 41 & 0.042 [0.030, 0.055]\\
Fixed-winner & $\Gy$ & $Z_0^{\mathrm{sel}}$ & $T_{\mathrm{fix}}$ & 999 & 17 & 0.018 [0.010, 0.027]\\
Fixed-winner, lock draws & $\Gy$ & $Z_0^{\mathrm{sel}}$ & Region, $\lambda{=}10$ & 1000 & 22 & 0.023\\
\bottomrule
\end{tabular}
\end{table}

\begin{figure}[H]
\centering
\begin{tikzpicture}[x=\textwidth/411.015,y=-\textwidth/411.015]
\node[anchor=north west,inner sep=0] at (0,0) {\includegraphics[width=\textwidth]{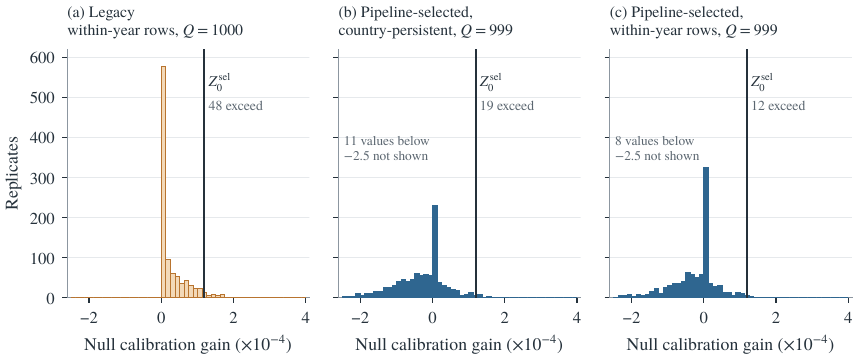}};
\node[anchor=base west,inner sep=0,text=rgInk,font=\fontfamily{ptm}\fontsize{9.3}{10}\selectfont] at (66.75,8.4) {(lock),};
\end{tikzpicture}
\caption{WDI null distributions for three constructions. The vertical line is $Z_0^{\mathrm{sel}}$, and the annotation is the number of null values $\ge Z_0^{\mathrm{sel}}$. The legacy null is bounded below by 0 because \textsf{NoContext} enters the maximum. Bin width $0.125\times10^{-4}$; 11 and 8 values below $-2.5\times10^{-4}$ in panels (b) and (c) are outside the plotted range.}
\label{fig:nulls}
\end{figure}

\textbf{Residual constructions.} With centered columns, $\mathbf1^\top\widetilde\Phi=0$, so the unpenalized intercept is $\hat b=\overline e_{\mathrm{tr}}$ for every family and $\lambda$. Table~\ref{tab:residextra} separates this offset from the category deviation. OOB predictions average the trees for which a row is out of bag (148--226 trees per row); regenerating the bootstrap indices from the stored per-tree seeds reproduces every leaf value of the first 50 trees to $1.1\times10^{-16}$. Forward residuals start from the fourth training year. With in-sample residuals the O3 pipeline reproduces the locked selection, calibration, block and test gains to machine precision.

\begin{table}[H]
\centering\small
\caption{Residual constructions (retrospective). $n$: training rows with a residual; Mean: mean residual, equal to the shared intercept $\hat b$ ($\times10^{-4}$); Int.: gains of the intercept-only correction $r\equiv\hat b$; Dev.\ test: test gain of the selected candidate with its intercept removed ($\times10^{-4}$).}
\label{tab:residextra}
\setlength{\tabcolsep}{4pt}\begin{tabular}{@{}llrrrrrr@{}}
\toprule
Origin & Res. & $n$ & RMSE & Mean & Int. cal. & Int. test & Dev. test\\
\midrule
O1 & In & 1551 & 0.0299 & $-1.5$ & $-0.06$ & $-0.18$ & $1.23$\\
O1 & OOB & 1551 & 0.0648 & $-3.7$ & $-0.15$ & $-0.46$ & $0.66$\\
O1 & Fwd & 974 & 0.0634 & $-46.4$ & $-3.15$ & $-7.29$ & --\\
O2 & In & 1943 & 0.0310 & $-0.7$ & $0.07$ & $0.05$ & $0.79$\\
O2 & OOB & 1943 & 0.0641 & $-1.9$ & $0.19$ & $0.13$ & $0.80$\\
O2 & Fwd & 1366 & 0.0645 & $-5.4$ & $0.52$ & $0.35$ & $-0.52$\\
O3 & In & 2336 & 0.0328 & $-2.0$ & $-0.16$ & $-0.26$ & $1.18$\\
O3 & OOB & 2336 & 0.0663 & $-5.7$ & $-0.47$ & $-0.74$ & $1.18$\\
O3 & Fwd & 1759 & 0.0671 & $-23.5$ & $-2.31$ & $-3.46$ & $-0.38$\\
O4 & In & 2734 & 0.0350 & $-1.2$ & $-0.15$ & $-0.23$ & $2.00$\\
O4 & OOB & 2734 & 0.0661 & $-4.1$ & $-0.54$ & $-0.79$ & $2.41$\\
O4 & Fwd & 2157 & 0.0674 & $-18.3$ & $-2.60$ & $-3.71$ & --\\
\bottomrule
\end{tabular}
\end{table}

\begin{table}[H]
\centering\small
\caption{RMSE of the forest $g$ against two forecasts that need no model: zero change and the training-period mean. Bold: lowest RMSE within each split and origin.}
\label{tab:naive}
\begin{tabular}{@{}lrrrrrrrrr@{}}
\toprule
 & \multicolumn{3}{c}{Selection} & \multicolumn{3}{c}{Calibration} & \multicolumn{3}{c}{Test}\\
\cmidrule(lr){2-4}\cmidrule(lr){5-7}\cmidrule(l){8-10}
Origin & $g$ & Zero & Mean & $g$ & Zero & Mean & $g$ & Zero & Mean\\
\midrule
O1 & 0.06535 & \textbf{0.06389} & 0.06540 & 0.07462 & 0.07584 & \textbf{0.07415} & \textbf{0.06415} & 0.06505 & 0.06455\\
O2 & 0.07452 & 0.07584 & \textbf{0.07428} & \textbf{0.07156} & 0.07311 & 0.07223 & 0.05536 & 0.05368 & \textbf{0.05340}\\
O3 & \textbf{0.06920} & 0.07311 & 0.07224 & 0.05610 & 0.05584 & \textbf{0.05556} & 0.05558 & \textbf{0.05551} & 0.05609\\
O4 & 0.05668 & 0.05584 & \textbf{0.05555} & 0.05268 & 0.05143 & \textbf{0.05129} & 0.05952 & \textbf{0.05807} & 0.05904\\
\bottomrule
\end{tabular}
\end{table}

\textbf{Zero-floor comparison on the 12 cells.} Holding the selected candidates fixed, we apply five rules: deploy whenever $\Dh_{\mathrm{cal}}>0$, floor only ($\Dh_{\mathrm{cal}}\ge3\times10^{-4}$), randomization only ($p_{\Gc}\le0.05$), block check only, and the full rule. The two \textsf{NoContext} cells always fall back. The floor-only and full rules agree cell by cell, so $v_i=0$ in all 12 cells.

\begin{table}[H]
\centering\small
\caption{Retrospective zero-floor comparison on the 12 WDI origin--residual cells (10 contextual winners). FD: deployed cells with negative test gain; miss: rejected contextual winners with positive test gain. Mean test gain averages $a\Dh_{\mathrm{te}}$ over all 12 cells ($\times10^{-4}$). The cells overlap in time, so no interval is reported. Bold: best value per column.}
\label{tab:wdiablate}
\setlength{\tabcolsep}{4pt}
\begin{tabular}{@{}lrrrr@{}}
\toprule
Policy & Deploy & FD & Miss & Mean test gain\\
\midrule
Positive calibration gain only & 5 & \textbf{1} & \textbf{4} & \textbf{0.248}\\
Floor only ($3\times10^{-4}$) & 2 & \textbf{1} & 7 & 0.064\\
Randomization only ($\Gc$) & 5 & \textbf{1} & \textbf{4} & \textbf{0.248}\\
Block check only & 3 & \textbf{1} & 6 & 0.134\\
Full rule & 2 & \textbf{1} & 7 & 0.064\\
\bottomrule
\end{tabular}
\end{table}

\section{Synthetic Protocol and Additional Results}\label{app:synthetic}
\textbf{Generative model.} For $N=96$ countries and $t=0,\dots,15$: Region $R_i\in\{1,\dots,4\}$ with probabilities $(.25,.25,.25,.25)$, $(.52,.24,.16,.08)$ or $(.73,.14,.09,.04)$; Income $I_i\in\{1,2,3\}$ with probabilities $(.35,.40,.25)$, independent of $R_i$; country effect $u_i\sim N(0,.34^2)$; covariates $x_{i,t}=\bar x_i+\eta_{i,t}+0.04t(1,-.5,.3,.2)$ with $\bar x_i\sim N(0,.75^2I_4)$, $\eta_{i,t}\sim N(0,.65^2I_4)$; AR(1) noise $\varepsilon_{i,t}=.47\varepsilon_{i,t-1}+\nu_{i,t}$, $\nu\sim N(0,.36^2)$; and $y_{i,t}=.75x_{i,t,1}-.5x_{i,t,2}+.24x_{i,t,3}+.08t/16+u_i+s\,\kappa_t\,\gamma_{R_i}+\varepsilon_{i,t}$, with $\gamma=(-1.2,-.35,.45,1.1)$ centered over countries. Drift $\kappa_t$: stable $1$; moderate $1$ for $t<7$, else $\max(.35,1-.055(t-6))$; calibration decay $1$ for $t<7$, else $\max(-.5,1-.14(t-6))$; test-only reversal $1$ for $t<13$ and $-1$ for $t\ge13$. The global model is Ridge on $x$ with intercept (slope penalty 2), fit on years 0--6. The synthetic Region$\times$Income family uses the 12 interaction cells, whereas the WDI family is additive.

\textbf{Null trials.} Among the 231 zero-signal trials that select a contextual candidate, the mean calibration gain is 0.0076 and the maximum 0.036; 79.7\% have positive test gain (mean 0.0074), a consequence of the aliasing between country effects and Region. Only 2 of the 240 pass the 0.03 floor, so in this design the floor, not the screen, is the binding guard.

\begin{table}[H]
\centering\small
\caption{Statistic-construction ablation on identical permutation streams ($Q=99$, $\alpha=0.05$): rejections out of 240 trials per signal strength, and the number of the 720 trials in which the screen decision differs from the pipeline statistic under the same group. The 9 trials that select \textsf{NoContext} count as non-rejections. Bold: highest power under $\Gc$ (all constructions tie at $s=1.10$); $\Gy$ rows are not ranked because that group is anti-conservative.}
\label{tab:ablation}
\begin{tabular}{@{}llrrrr@{}}
\toprule
Group & Construction & $s=0$ & $s=0.45$ & $s=1.10$ & Differs from pipeline\\
\midrule
$\Gc$ & Pipeline-selected & 10 & 172 & 180 & --\\
 & Fixed-winner & 12 & 173 & 180 & 3\\
 & Calibration-max & 12 & \textbf{176} & 180 & 6\\
 & Legacy-style & 10 & 172 & 180 & 0\\
\midrule
$\Gy$ & Pipeline-selected & 178 & 186 & 181 & --\\
 & Fixed-winner & 168 & 185 & 181 & 13\\
 & Calibration-max & 178 & 201 & 189 & 45\\
 & Legacy-style & 160 & 185 & 181 & 19\\
\bottomrule
\end{tabular}
\end{table}

\begin{table}[H]
\centering\small
\caption{Harmful deployments by drift mechanism (180 trials per column). Bold: lowest count per column.}
\label{tab:fddrift}
\setlength{\tabcolsep}{5pt}
\begin{tabular}{@{}lrrrr@{}}
\toprule
Policy & Stable & Moderate & Calibration decay & Test-only reversal\\
\midrule
Always selected & 5 & 29 & 133 & 136\\
Calibration threshold & \textbf{0} & 12 & 1 & 120\\
Country-cluster LCB & \textbf{0} & \textbf{10} & \textbf{0} & \textbf{119}\\
Randomization only & \textbf{0} & 12 & 1 & 120\\
Temporal only & 3 & 16 & 7 & 125\\
Full RAEGL & \textbf{0} & 12 & 1 & 120\\
\bottomrule
\end{tabular}
\end{table}

\begin{table}[H]
\centering\small
\caption{All policies over 720 trials. Selection-only coincides with always-selected, and the calibration threshold with floor-only. FD cond.: harmful deployments per deployment. Bold: best value per column; deployment count has no preferred direction.}
\label{tab:allpolicies}
\begin{tabular}{@{}lrrrrrr@{}}
\toprule
Policy & Deploy & FD & FD cond. & Miss & Mean gain & Mean regret\\
\midrule
Always selected / selection only & 711 & 303 & 42.6\% & \textbf{0} & $-0.0617$ & 0.1197\\
Calibration threshold / floor only & 356 & 133 & 37.4\% & 185 & $-0.0241$ & 0.0821\\
Country-cluster LCB & 373 & \textbf{129} & 34.6\% & 164 & $-0.0233$ & 0.0813\\
Randomization only & 362 & 133 & 36.7\% & 179 & $-0.0240$ & 0.0820\\
Temporal only & 492 & 151 & \textbf{30.7\%} & 67 & $\mathbf{-0.0224}$ & \textbf{0.0804}\\
Randomization + floor & 354 & 133 & 37.6\% & 187 & $-0.0242$ & 0.0822\\
Randomization + temporal & 362 & 133 & 36.7\% & 179 & $-0.0240$ & 0.0820\\
Full RAEGL & 354 & 133 & 37.6\% & 187 & $-0.0242$ & 0.0822\\
\bottomrule
\end{tabular}
\end{table}

\begin{table}[H]
\centering\small
\caption{Post hoc sensitivity of the full rule to $(\alpha,\eprac)$ over 720 trials ($\Gc$ $p$-values; block check fixed). Not used to choose any reported setting. With $\eprac=0$ the screen and block check remain active, so these rows do not describe a gain-only rule. Bold: best value per column.}
\label{tab:thresh}
\begin{tabular}{@{}rrrrrr@{}}
\toprule
$\alpha$ & $\eprac$ & Deploy & FD (uncond.) & Mean gain & Mean regret\\
\midrule
0.01 & 0.00 & 46.8\% & 17.8\% & $-0.0244$ & 0.0824\\
0.01 & 0.01 & 46.8\% & 17.8\% & $-0.0244$ & 0.0824\\
0.01 & 0.03 & 46.8\% & 17.8\% & $-0.0244$ & 0.0824\\
0.01 & 0.06 & 44.0\% & 17.1\% & $-0.0242$ & 0.0822\\
0.01 & 0.10 & 33.9\% & \textbf{13.2\%} & $\mathbf{-0.0206}$ & \textbf{0.0786}\\
0.05 & 0.00 & 50.3\% & 18.5\% & $-0.0240$ & 0.0820\\
0.05 & 0.01 & 50.3\% & 18.5\% & $-0.0240$ & 0.0820\\
0.05 & 0.03 & 49.2\% & 18.5\% & $-0.0242$ & 0.0822\\
0.05 & 0.06 & 44.2\% & 17.1\% & $-0.0241$ & 0.0821\\
0.05 & 0.10 & 33.9\% & \textbf{13.2\%} & $\mathbf{-0.0206}$ & \textbf{0.0786}\\
0.10 & 0.00 & 51.9\% & 18.8\% & $-0.0238$ & 0.0818\\
0.10 & 0.01 & 51.9\% & 18.8\% & $-0.0238$ & 0.0818\\
0.10 & 0.03 & 49.3\% & 18.5\% & $-0.0242$ & 0.0822\\
0.10 & 0.06 & 44.2\% & 17.1\% & $-0.0241$ & 0.0821\\
0.10 & 0.10 & 33.9\% & \textbf{13.2\%} & $\mathbf{-0.0206}$ & \textbf{0.0786}\\
\bottomrule
\end{tabular}
\end{table}

\begin{figure}[H]
\centering
\includegraphics[width=\textwidth]{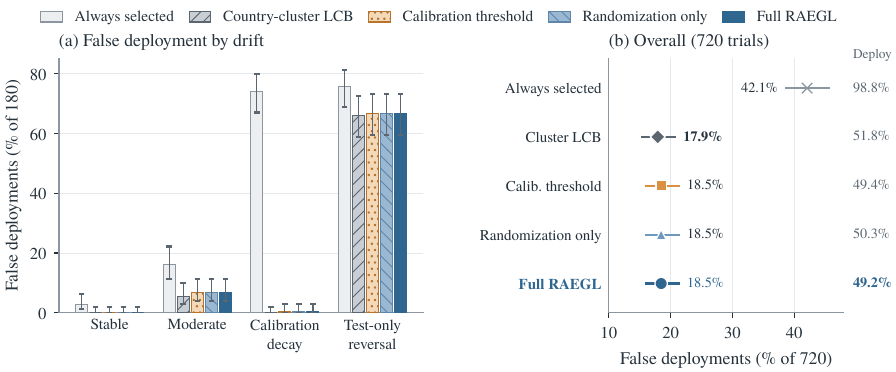}
\caption{\textbf{Policies across drift mechanisms.} (a) Unconditional harmful-deployment frequency per mechanism (180 trials each; Wilson 95\% intervals). (b) Overall frequency and deployment rate. Single operating points, not risk--coverage curves; the four gated policies overlap within Monte Carlo error.}
\label{fig:policies}
\end{figure}

\textbf{Drift regimes.} Under stable context RAEGL deploys in all 120 trials with $s>0$, in none of the 60 with $s=0$, and never deploys harmfully; in 55 of those 60 zero-signal trials the rejected candidate still improved test RMSE slightly through aliasing, which shows that the label null and test usefulness are different targets. Under moderate decay RAEGL deploys harmfully 12 times (always-selected: 29). Under calibration decay the weakening is visible in $\Dcal$ and the gated rules almost never deploy. Under test-only reversal every calibration-based rule deploys in essentially all 120 trials with $s>0$, each harmfully (Fig.~\ref{fig:policies}, Table~\ref{tab:drift}).

\begin{table}[H]
\centering\small
\caption{RAEGL deployments by drift and signal (of 60 each), and zero-signal rejections of the screen by group (of 60 each).}\label{tab:drift}
\begin{tabular}{@{}lrrrrr@{}}
\toprule
Drift & $s=0$ & $s=0.45$ & $s=1.10$ & $\Gc$ rejections ($s{=}0$) & $\Gy$ rejections ($s{=}0$)\\
\midrule
Stable & 0 & 60 & 60 & 3 & 52\\
Moderate & 1 & 51 & 60 & 2 & 39\\
Calibration decay & 0 & 1 & 0 & 2 & 41\\
Test-only reversal & 1 & 60 & 60 & 3 & 46\\
\bottomrule
\end{tabular}
\end{table}

\section{Specifying the Protocol for a New Panel}\label{app:design}
The WDI settings are application choices. For a new panel we suggest declaring the following before calibration data are examined.
\textbf{Baseline and horizon.} Fix the deployment horizon and the global model, and compare the global model with simple domain forecasts on training and selection data; a component gate cannot make a weak baseline worth deploying. Admit only contexts observable at forecast time with adequate group support, and freeze the grid and tie-breaks.
\textbf{Phases.} Use non-overlapping training, selection, calibration and evaluation periods, with a gap where reporting latency requires one. The calibration period should span the temporal variation the application expects; with few blocks, a per-block sign check is a weak stability signal.
\textbf{Floor.} State $\eprac$ in RMSE units from an external minimum useful effect, or as $\eta R_{\mathrm{ref}}$ with $\eta$ and a pre-calibration reference risk $R_{\mathrm{ref}}$ fixed in advance. A zero floor, a cost-based floor and a confidence-bound criterion express different preferences.
\textbf{Screen.} Use a permutation group only if its label symmetry has a randomization or scientific justification; otherwise report the $p$-value as a diagnostic. Fix $Q$ in advance and report Monte Carlo error near the boundary.
\textbf{Reporting and reapproval.} Report deployment rate, unconditional and conditional harm, misses, mean gain and regret. An approval covers one declared horizon; a new horizon requires a newly specified run, and repeated approval has no sequential error guarantee.

\section{Data and Code Availability}\label{app:repro}
The supplement contains the WDI source tables and processed panel, the frozen O3 models, lock record and prediction-level outputs, the notebooks that build the panel and run the locked experiment, the code of the residual and rolling-origin study with its 12 output records, and scripts that recompute the post-decision diagnostics and the nested-veto identity. For the synthetic study the supplement contains the aggregate counts behind Tables~\ref{tab:policies} and \ref{tab:ablation}--\ref{tab:drift}; the generating script and per-trial decisions were not retained, so these tables can be checked against the stored aggregates but not regenerated. The protocol configuration file lists 200 null replicates, whereas the lock record and null outputs contain 1000; we treat the lock record as authoritative.

\end{document}